\documentclass[lettersize,journal]{IEEEtran}
\usepackage{amsmath,amsfonts}
\usepackage{algorithmic}
\usepackage{algorithm}
\usepackage{array}
\usepackage[caption=false,font=normalsize,labelfont=sf,textfont=sf]{subfig}
\usepackage{textcomp}
\usepackage{stfloats}
\usepackage{url}
\usepackage{verbatim}
\usepackage{graphicx}
\usepackage{multirow}
\usepackage{cite}

\usepackage{booktabs}       

\usepackage{amssymb}

\begin{document}

\title{Learning Spiking Neural Network from Easy to Hard task}

\author{Lingling Tang, Jiangtao Hu, Hua Yu, Surui Liu, Jielei Chu}




\maketitle

\begin{abstract}
Starting with small and simple concepts, and gradually introducing complex and difficult concepts is the natural process of human learning. Spiking Neural Networks (SNNs) aim to mimic the way humans process information, but current SNNs models treat all samples equally, which does not align with the principles of human learning and overlooks the biological plausibility of SNNs. To address this, we propose a CL-SNN model that introduces Curriculum Learning(CL) into SNNs, making SNNs learn more like humans and providing higher biological interpretability. CL is a training strategy that advocates presenting easier data to models before gradually introducing more challenging data, mimicking the human learning process. We use a confidence-aware loss to measure and process the samples with different difficulty levels. By learning the confidence of different samples, the model reduces the contribution of difficult samples to parameter optimization automatically. We conducted experiments on static image datasets MNIST, Fashion-MNIST, CIFAR10, and neuromorphic datasets N-MNIST, CIFAR10-DVS, DVS-Gesture. The results are promising. To our best knowledge, this is the first proposal to enhance the biologically plausibility of SNNs by introducing CL. 
\end{abstract}


\section{Introduction}
\IEEEPARstart{E}{fficient} and low-power earning and information processing, similar to the human brain has always been a goal pursued in the field of deep learning. The design of Spiking Neural Networks (SNNs) is inspired by the neuroscience research on human brain. It attempts to simulate the behavior and information processing of neurons to better approximate the functioning principles of the biological brain\cite{jang2019introduction, maass1997networks}. Unlike traditional Artificial Neural Networks (ANNs)\cite{chu2018restricted, chu2022micro, chu2020unsupervised}, which use continuous-valued activations, SNNs use discrete, time-based representations spikes or action potentials to transmit and process information\cite{yao2023attention}. SNNs have the potential for efficient and low-energy information processing, as they only transmit spikes when necessary. This makes SNNs attractive for implementing neuromorphic hardware and energy-efficient computing systems\cite{roy2019towards}. SNNs can capture the temporal dynamics and precise timing information of neural computations, which are crucial for areas such as speech recognition\cite{han2023complex}, event-based processing\cite{roy2019towards}, and sensory integration\cite{jia2022motif}.

Unsupervised SNNs\cite{hazan2018unsupervised, xiang2019stdp, paredes2019unsupervised} are primarily based on biological principles, with the most common being the Spiking Time Dependency Plasticity (STDP) rule\cite{han2021survey}. 
It adjusts the synaptic weights based on the timing relationship between pre-synaptic and post-synaptic neurons' spike emissions, demonstrating high biological plausibility. However, it is challenging to achieve good performance on deep networks.

Supervised learning methods\cite{mostafa2017supervised, sporea2013supervised}, on the other hand, train SNNs using labeled data by minimizing the error between predicted outputs and true labels to 
adjust network parameters\cite{dampfhoffer2023backpropagation}. Since spiking neural networks transmit discrete spike signals and lack differentiability\cite{huh2018gradient}, surrogate gradient\cite{neftci2019surrogate} methods are commonly used for backpropagation to optimize the parameters. This approach has shown promising results and provides a relatively simple solution for handling the non-differentiability of SNNs.Another learning approach is ANN to SNN conversion, where an ANNs is first trained, and then a structurally equivalent SNNs is constructed, initializing the SNN's weights with those of the ANNs. This conversion allows SNNs to achieve performance close to that of ANNs\cite{stockl2021optimized, li2021free, deng2021optimal}. However, it requires longer time steps\cite{wu2021progressive} and is not suitable for event-based datasets.

When we learn new knowledge, we often start with small and simple concepts and gradually introduce complex and challenging knowledge. This is a natural learning process for us. Just as we receive education based on a curriculum designed by schools to continuously enhance our knowledge. SNNs is proposed to mimic the way humans process information\cite{roy2019towards}. However, the current SNNs models treat all samples equally during the training process without considering whether the model's learning capacity can effectively acquire relatively complex and challenging knowledge\cite{fang2021incorporating, li2021differentiable, fang2021deep}. This approach is not reasonable and does not align with the natural laws of human learning, resulting in lower biological interpretability. Therefore, to improve the biological plausibility of SNNs, we propose introducing Curriculum Learning (CL) into SNNs to make them learn more like humans.

CL is inspired by the process of human learning new knowledge. Its main idea is to gradually increase the training difficulty, allowing the model to learn useful knowledge more quickly and better generalize to new data in later stages of training\cite{bengio2009curriculum}. Specifically, CL provides training data to the model in a certain order called a "curriculum"\cite{dillon2009questions}. Initially, the model is exposed to simple training samples, such as easily classifiable samples or correctly labeled ones. Once the model performs well, the difficulty is gradually increased by providing more complex and challenging samples. This process aligns closely with the process of human learning new knowledge.

The sequential learning process of CL mimics the natural way humans acquire new knowledge. In the\cite{wang2021survey}, CL is summarized as a combination of a difficulty estimator and a training scheduler. The difficulty estimator evaluates the difficulty of samples based on selected evaluation features such as complexity\cite{weinshall2018curriculum}, noise\cite{choi2019pseudo, guo2018curriculumnet}, or suggestions from a mature teacher network\cite{weinshall2018curriculum}. The training scheduler then feeds the samples to the model in the order of difficulty according to the designed scheduling rules, by which can improve the model convergence speed. Pre-defined CL, as described in\cite{wang2021survey}, relies on manually defining methods for evaluating sample difficulty and scheduling strategies. While simpler to implement, this approach tends to overlook the model's feedback during the learning process. Moreover, it is challenging to determine the most suitable training schedule, such as when and how much to introduce more challenging samples, based solely on prior knowledge, especially for specific tasks and datasets. It is more suitable for small datasets. In contrast to pre-defined CL, automatic CL introduces the concept of a teacher network. The teacher network can be the student network itself, which evaluates its own learning progress based on the training loss and adjusts the training samples accordingly. Alternatively, it can be a well-trained, more advanced network model that assesses sample difficulty based on the teacher model's performance on those samples and dynamically adjusts the sample inputs based on feedback from the student network model.

Confidence-aware loss has been proposed as a type of loss function that considers the model's confidence or certainty in its predictions\cite{kendall2018multi, novotny2018self}. During the training process, it aims to assign higher importance to samples with higher confidence and lower importance to samples with lower confidence. This is achieved by introducing sample weights, namely confidence, and adjusting the weights of samples during the training process to modulate their contribution to the model parameter updates during backpropagation. The confidence-aware loss achieves curriculum learning without the need for predefined curriculum design.

The paper's main contributions can be summarized as follows:
\begin{itemize}
\item {We propose a CL-SNN model that has higher biological interpretability by introducing CL into SNNs. The model dynamically evaluates the difficulty of samples, assigns higher confidence to simple samples, and amplifies their contribution in backpropagation. It automatically reduces the impact of more difficult samples on parameter updates. This approach exhibits high biological plausibility and effectively simulates the process of human learning new knowledge. To our best knowledge, this is the first attempt to enhance the biologically plausibility of SNNs by introducing CL.}

\item {We evaluate our CL-SNN model on three static image datasets, MNIST, CIFAR10 and Fashion-MNIST, as well as three neuromorphic datasets, including DVS-Gesture, N-MNIST, and CIFAR10-DVS for classification tasks. The results surpass the current state-of-the-art experimental results for all six datasets.}
\end{itemize}

The rest of the paper is organized as follows. Section \uppercase\expandafter{\romannumeral2} is the proposed methods. Section \uppercase\expandafter{\romannumeral3} introduces the experiment results. And in section \uppercase\expandafter{\romannumeral4} we make a summary and disscussion.

\begin{figure*}[!htbp]
\vspace{0.5mm} \centering

\begin{tabular}{cc}
    \includegraphics[width=0.9\linewidth]{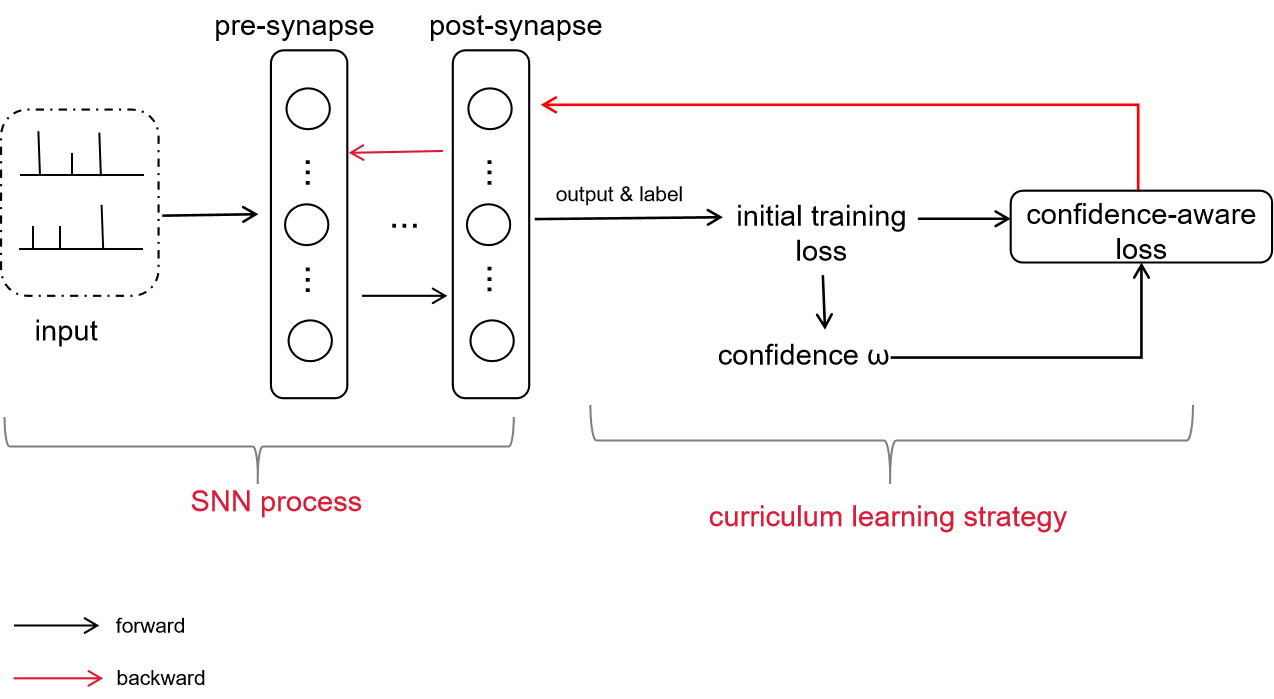}
\end{tabular}
 \caption{The structure of proposed CL-SNN.}\label{fig:cl-snn}
\end{figure*}

\section{Methods}
\subsection{Neuron model}
The most commonly used spiking neuron model for SNNs is Leaky Integrate and Fire (LIF) neuron model\cite{gerstner2014neuronal}. It features a simple computation process while preserving essential biological
characteristics. The dynamics of LIF can be described as follows:

\begin{equation}
\begin{aligned}
\tau_m \frac{\mathrm{d} V\left(t\right)}{\mathrm{d} t} = -\left(V\left( t \right) -V_{r} \right) +  I\left(t\right),
\end{aligned}
\end{equation}

in which $V_{r}$ is the fixed resting potential, $V(t)$ represents the membrane potential at time-step $t$. $V(t)$ decays continuously and, in the absence of input, it will decay until it reaches $V_{r}$. $\tau_m$ represents the membrane time constant. The LIF neuron model combines biological plausibility with computational convenience. Since SNNs transmit discrete spike signals, the dynamical mechanism of SNNs can be described as follows:

\begin{equation}
\begin{aligned}
Q_t = f( V_{t-1}, I_t),\label{Q_t}
\end{aligned}
\end{equation}

\begin{equation}
\begin{aligned}
O_t = \theta( Q_t - V_{th}),\label{S_t}
\end{aligned}
\end{equation}

\begin{equation}
\begin{aligned}{}
V_t = Q_t\left(1-O_t\right) + V_{r}O_t.\label{V_t}
\end{aligned}
\end{equation}

where $Q_t$ represents the membrane potential after receiving input, and depend on the membrane potential at time-step $t-1$ and the input $I_t$. $I_t=\textbf{w}_{ji}*O_t$ and $\textbf{w}_{ji}$ denotes the weight for spiking neuron $i$ and $j$. $Q_t$ is calculated by the function $f(x)$. $O_t$ represents the output of the pre-neuron,in which
\begin{equation}
\begin{aligned}
\theta(x) =
\left\{ 
    \begin{array}{lc}
        1 ,& x \geqslant 0 \\
        0 ,& x<0\\
    \end{array}
\right.
\end{aligned}
\end{equation}

$O_t=1$ means the potential after receiving input reaches the threshold $V_{th}$ and generate a spike, and $O_t=0$ means neuron keep silent. $V_t$ represents the membrane potential after fire process. Eqs.\eqref{Q_t}\eqref{S_t}\eqref{V_t} describe the charging, discharging, and resetting processes of a spiking neuron, respectively. The differences between different spiking neuron models mainly lie in the function $f(x)$. The $f(x)$
for LIF neuron is defind as:
\begin{equation}
\begin{aligned}
Q_t = f( V_{t-1}, I_t) =V_{t-1} + \frac{1}{\tau_m}\left(- \left(V_{t-1} - V_{r}\right)+I_t\right) \label{Q_t1}.
\end{aligned}
\end{equation}
Most SNNs models that utilize LIF neurons typically set the membrane time constant $\tau_m$ as a
fixed constant. However, in \cite{fang2021incorporating} proposed the importance and necessity of a learnable $\tau_m$ on model performance. They introduced the PLIF model, which simultaneously learns $\tau_m$ and synaptic weights during training, resulting in improved fitting capabilities. To avoid errors caused by $\tau_m$ in the denominator during the learning process, Eq.\eqref{Q_t1} is rewritten as follows:
\begin{equation}
\begin{aligned}
Q_t = \left(1 - g(a)\right)V_{t-1} + g(a)\left(V_{r} + I_t\right) \label{Q_t2}
\end{aligned}
\end{equation}
where $\tau_m = \frac{1}{g(a)} \in (1, + \infty)$, a is a learnable parameter. In this paper, $g(a)$ is taken as the sigmoid function, $g(a)=\frac{1}{1+e^{(-a)}}$.
\subsection{Surrogate gradient}
Because SNNs transmit discrete spike signals and exhibit non-continuous neuron responses. And $\theta (x)$ is non-differentiable,the derivative of $\theta (x)$ is: 
\begin{equation}
\begin{aligned}
\theta'(x) =
\left\{ 
    \begin{array}{lc}
        + \infty ,& x = 0 \\
        0 ,& x \neq 0\\
    \end{array}
\right.
\end{aligned}
\end{equation}
it becomes challenging to use traditional backpropagation for gradient updates and parameter
optimization in SNNs. Previous research\cite{neftci2019surrogate} has proposed the surrogate gradient method to enable backpropagation in SNNs. The principle of surrogate gradient is to use $y = \theta(x)$ during forward propagation and $\frac{\mathrm{d}y}{\mathrm{d}x} = \phi'(x)$ during backward  propagation, instead of $\frac{\mathrm{d}y}{\mathrm{d}x} = \theta'(x)$, where $\phi(x)$ represents the surrogate function. $\phi(x)$ is typically a smooth and continuous
function that has a similar shape to $\theta(x)$.
The training process of SNNs using the surrogate gradient method is as follows:
\begin{enumerate}
\item Forward Propagation: The input is passed through the network, simulating the spiking behavior of neurons, and generating spike outputs.
\item Surrogate Function Calculation: Based on the spike outputs, the value of the surrogate function is computed, and the gradient of the surrogate function is calculated.
\item Backpropagation: The gradients of the surrogate function are used for backpropagation to
update the network parameters.
\end{enumerate} 
The surrogate function used in this paper is:
\begin{equation}
\begin{aligned}
\phi(x) = \frac{1}{\pi} arctan(\frac{\pi}{2}\alpha x)+\frac{1}{2}.
\end{aligned}
\end{equation}
And we get:
\begin{equation}
\begin{aligned}
\frac{\mathrm{d}}{\mathrm{d}x}\phi(x) = \frac{\alpha}{2\left(1+\left(\frac{\pi}{2}\alpha x\right)^2\right)}.
\end{aligned}
\end{equation}
Gradient-based surrogate simplifies the computation of SNNs and achieves promising results.
\subsection{Curriculum learning with confidence-aware loss}
The main idea of CL is to accelerate the learning of useful knowledge by
gradually increasing the training difficulty, enabling the model to generalize better to new data in later stages of training. Common methods in CL involve presenting the model with simple data first and gradually introducing more challenging data in a predefined order of difficulty. This has led to the development of various difficulty measures and training schedulers\cite{wang2021survey}. However, most of these methods are suit for small datasets or focus on curriculum design at the data scheduling level. Moreover, it always requires additional resources for sample difficulty evaluation and sample scheduling. 

Previous research\cite{kendall2018multi, novotny2018self} has introduced the concept of confidence-aware loss, which is a type of loss function that takes into account the confidence or certainty of model predictions. During training, it aims to assign higher importance to samples with higher confidence and lower importance to samples with lower confidence. A confidence-aware loss function, denoted as $l(\hat{y}, y, \omega)$, introduces a learnable parameter $\omega$ as an additional input compared to the traditional loss function $l(\hat{y}, y)$, $\omega$ is the confidence or reliability of the current prediction result $\hat{y}$.

In \cite{Castellsastells2020superloss}, a general and lightweight approach to implementing the critical purpose of curriculum learning was proposed. It is based on confidence-aware loss. And its mathematical definition is as follows:
\begin{equation}
\begin{aligned}
CAL_\lambda\left(l_i, \omega_i\right) = \left(l_i - \epsilon \right) \omega_i +\lambda\left(log(\omega_i)\right)^2\label{L_},
\end{aligned}
\end{equation}
where $l_i$ represents the initial training loss, which can be calculated using common loss functions such as cross-entropy loss or mean squared error (MSE). $\omega_i$ is the confidence or certainty associated with sample $i$ and is a learnable parameter. $\epsilon$ is the threshold used to differentiate between easy and difficult samples and can be set as the average of the batch initial training loss $l_i$ or a predetermined constant. $\lambda$ is a hyper parameter that controls the regularization term. By dynamically learning the confidence $\omega$ for each sample, aiming to reduce the impact of difficult samples on parameter updates while amplifying the confidence of simple and reliable samples, thereby expanding their contribution to the model. As training progresses, the fitting
capacity of model continuously improves, leading to high confidence for all samples in the end. For the confidence $\omega_i$, it scales the learning level of the samples. To simplify the computation, the
confidence $\omega$ directly obtained from the training loss $l_i$ of the sample is defined as follows\cite{Castellsastells2020superloss}:
\begin{equation}
\begin{aligned}
\omega_i (l_i) = exp\left(-W\left(\frac{1}{2}max\left(-\frac{2}{e}, \eta\right)\right)\right)\label{l_i},
\end{aligned}
\end{equation}
with $\eta = \frac{l_i-\epsilon}{\lambda}$, and W is the Lambert W function. The final loss used for backpropagation is computed by combining the confidence $\omega$ obtained from the initial training loss $l_i$ and $l_i$ itself. In CL-SNN, we directly calculate the initial loss $l_i$ using the cross-entropy loss as follows:
\begin{equation}
\begin{aligned}
L = \frac{1}{N} \sum_i L_i = -\frac{1}{N}\sum_i\sum_{c=1}^M y_{ic}log(p_{ic})\label{L},
\end{aligned}
\end{equation}
Combining Eqs.\eqref{L_}\eqref{l_i}\eqref{L}, we obtain the confidence-aware loss function for CL-SNN. For
samples with a large initial training loss, they can be considered as more challenging for the
current model. Hence, we assign them lower confidence values to reduce their impact during
parameter updates. Conversely, for samples with smaller $l_i$ values, we can regard them as simpler and more reliable, this just like a dynamic CL.

\section{Experiments}

We evaluated the proposed CL-SNN model on classification tasks using six datasets, including
three static datasets (MNIST, CIFAR10, Fashion-MNIST) and three neuromorphic datasets
(N-MNIST, CIFAR10-DVS, DVS-Gesture).

\subsection{Experiment setting}

The experiments were conducted using a network architecture similar to that described in \cite{fang2021incorporating}. The SpikingJelly framework\cite{SpikingJelly} was utilized, without employing operations similar to Poisson encoding on the input, but instead directly feeding it into the network. For the MNIST, Fashion-MNIST, and N-MNIST datasets, we employed the same network architecture:
128c3-BN-MP2-128c3-BN-MP2-DP-FC2048-DP-FC100-AP10. Here, "128c3" refers to a convolutional layer with 128 channels and kernel size 3. "MP2" represents max-pooling layer with kernel size 2. "BN" represents batch normalization. "DP" refers to a dropout layer. We use the PLIF neuron described in Section 3.1. The detailed network architecture for other datasets are described in the code.

In the surrogate function, we set $\alpha=2$, so 
\begin{equation}
\begin{aligned}
\phi(x) = \frac{1}{\pi}arctan(\pi x) + \frac{1}{2},
\end{aligned}
\end{equation}
and $\phi' \left(x\right) = \frac{1}{1+\left(\pi x\right)^2}$.

For the threshold $\epsilon$ used to distinguish between easy and hard samples in the confidence-aware loss, there are two methods to define. One approach is to set $\epsilon$ as a fixed constant, while the other approach involves using a dynamic threshold. We conducted experiments for both cases.
In the first case, we set $\epsilon = log(c)$, where $c$ represents the number of classes. For datasets such as MNIST, CIFAR10, Fashion-MNIST, N-MNIST, and CIFAR10-DVS, $c$ is equal to $10$. For the DVS-Gesture dataset, $c$ is $11$, and we set $\lambda=0$. In the second case, we set $\epsilon$ as the average initial loss of each batch, resulting in a dynamic threshold, and in this case we set $\lambda=1$.

\subsection{Experiment results}
\begin{figure*}[!t]
\centering
\subfloat[]{\includegraphics[width=2.5in]{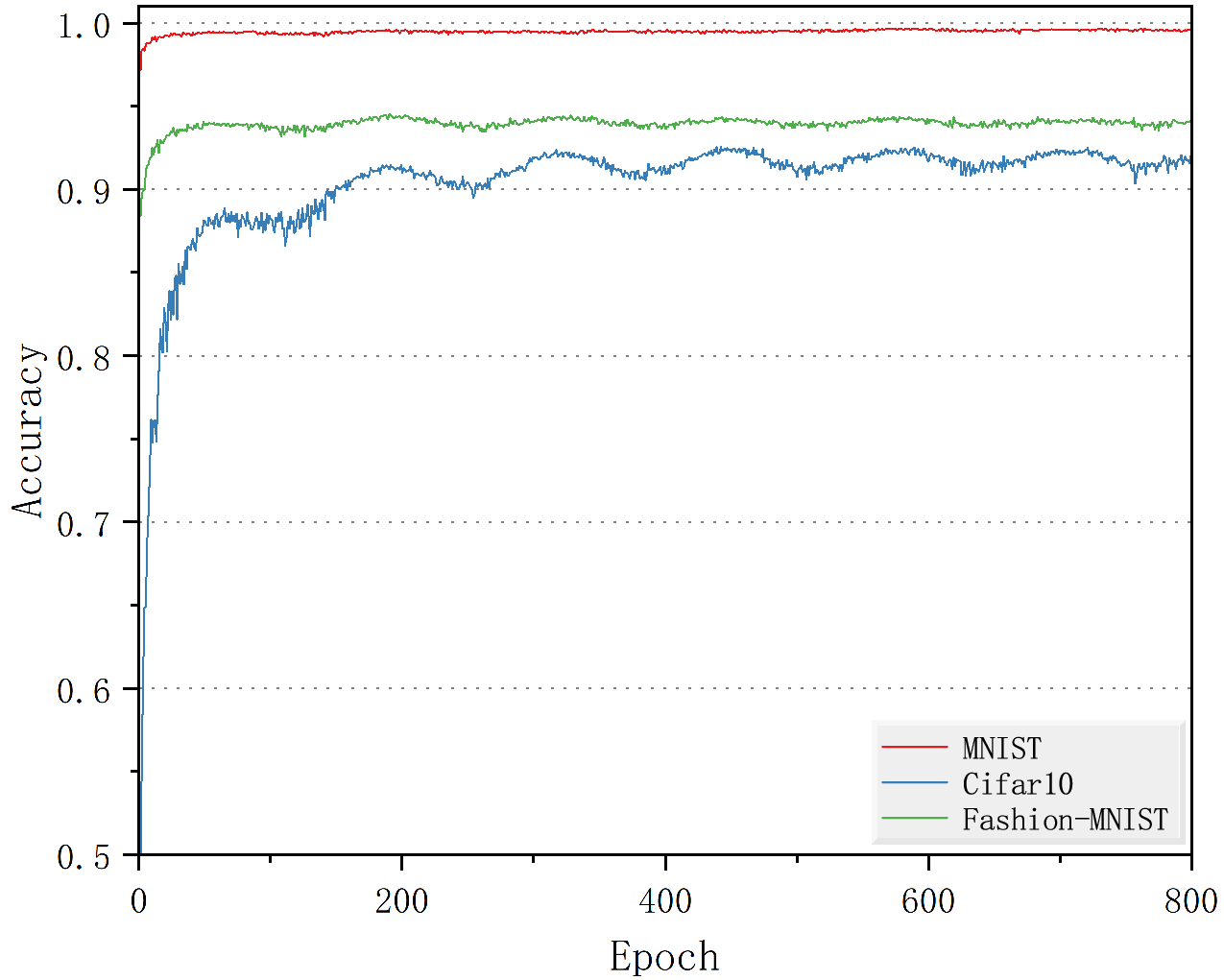}%
\label{fig:acc1}}
\hfil
\subfloat[]{\includegraphics[width=2.5in]{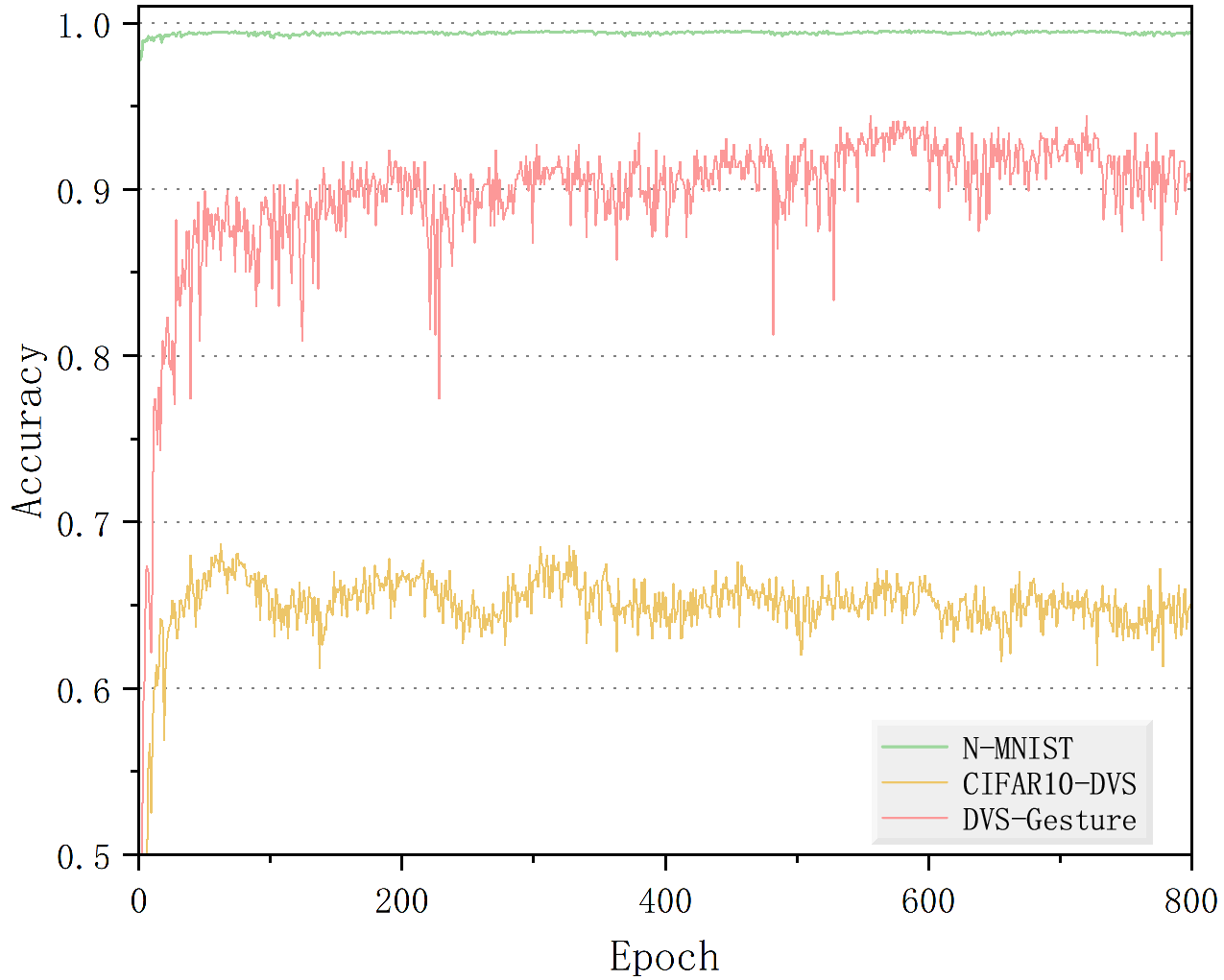}%
\label{fig:acc2}}
\caption{(a) Accuracy on static datasets. (b) Accuracy on neuromorphic datasets. }
\label{fig_acc}
\end{figure*}

The experimental results of our proposed CL-SNN model on static datasets for classification tasks, along with the comparison with state-of-the-art methods, are shown in the Table\eqref{tab:results1}. Our method outperforms the comparison methods on all six datasets, and our model exhibits higher biological plausibility, aligning with the principles of human knowledge acquisition. The results on neuromorphic datasets are show in Table\eqref{tab:results2}, all show a better performance. The accuracy change curve of the model for diffierent datasets is shown in Figure\eqref{fig_acc}.

In terms of the difficulty differentiation threshold $\epsilon$, experimental results show that dynamic $\epsilon$ performs better on CIFAR10, N-MNIST and DVS-Geature datasets. And for MNIST, Fashion-MNIST and CIFAR10-DVS, the fixed $\epsilon$ achieves better performance.

\begin{figure*}[!t]
\centering
\subfloat[]{\includegraphics[width=2.5in]{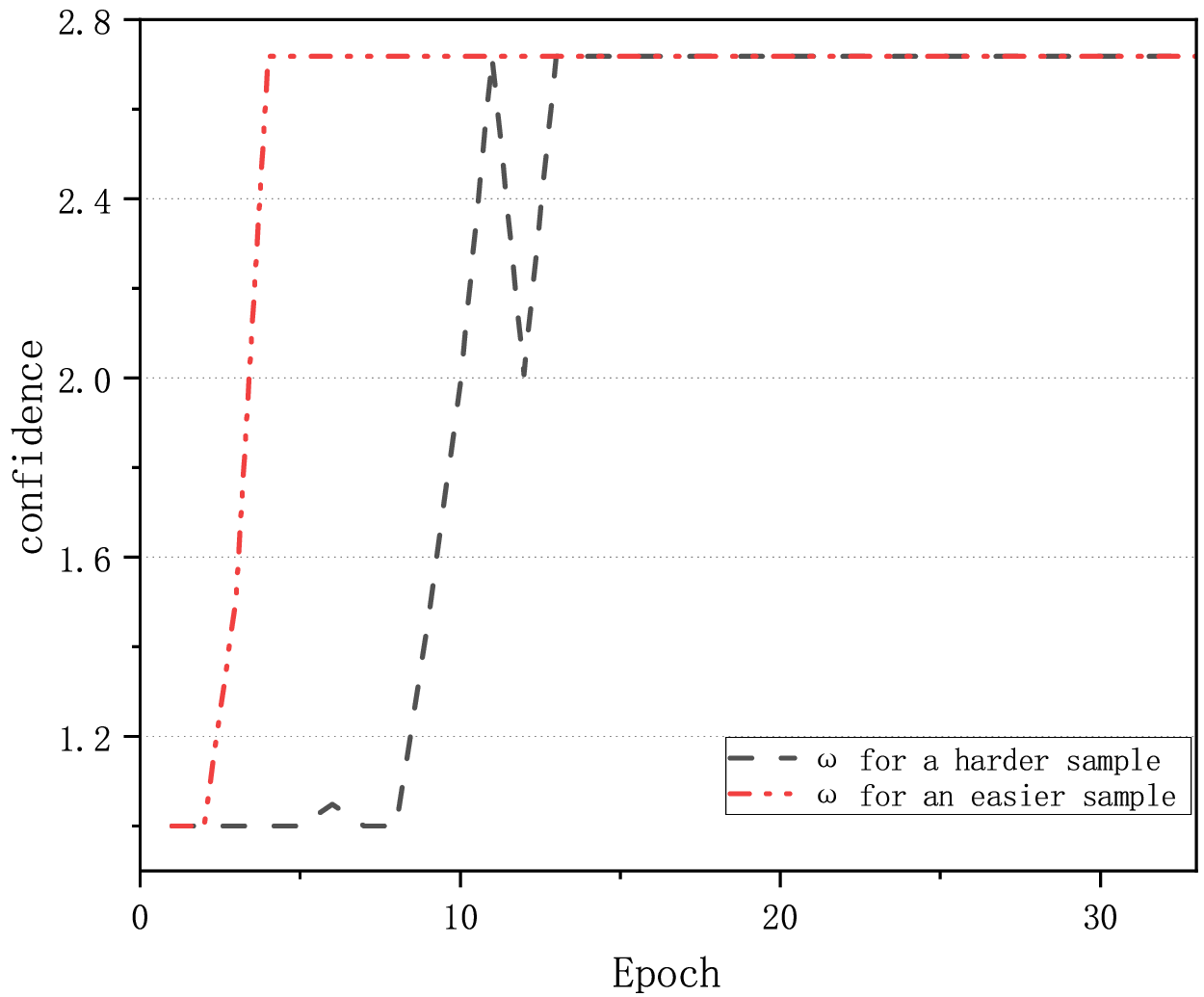}%
\label{fig:confidence}}
\hfil
\subfloat[]{\includegraphics[width=2.5in]{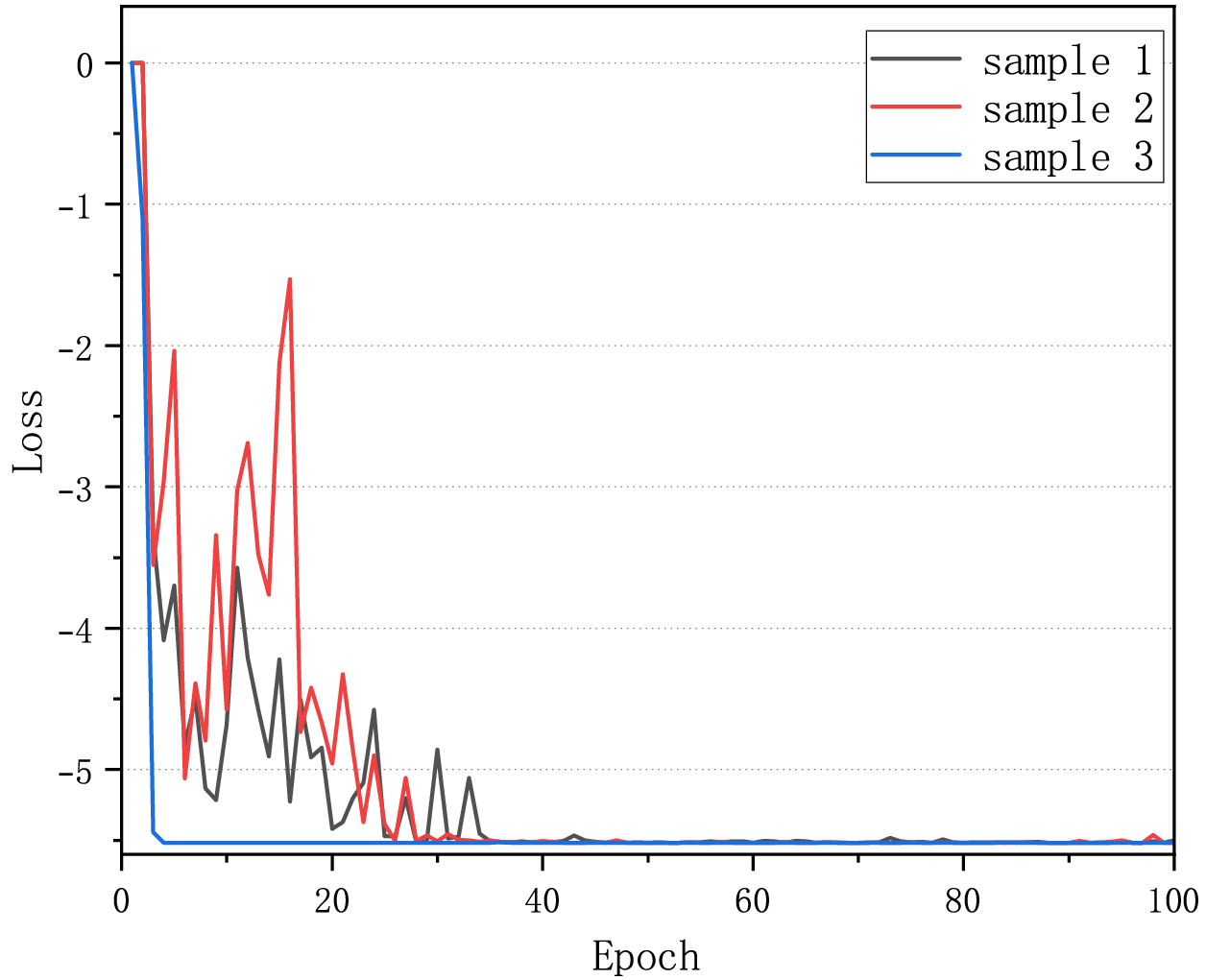}%
\label{fig:loss}}
\caption{(a) Confidence $\omega$ curve for diffierent samples. (b) Confidence-aware loss curve for diffierent samples.}
\label{figsim}
\end{figure*}

The confidence levels $\omega$ vary for samples of different difficulties, as shown in Figure\eqref{fig:confidence}. Easier samples reach the maximum confidence earlier, while more difficult samples take longer to reach maximum confidence. The level of confidence-aware loss for different samples is illustrated in Figure\eqref{fig:loss}.

\begin{table*}
\begin{center}
\caption{Comparisions of accuracy with other methods on static datasets. }
\label{tab:results1} \scalebox{1.0}{
\begin{tabular}{cccc}
\specialrule{0.1em}{5pt}{3pt} 
\textsf{ \bf{ Dataset }} & \textbf{ model   } & \textbf{ method   } & \textbf{ accuracy     }  \\
\specialrule{0.1em}{3pt}{0pt}\\
  \textsf{  } & {SCNN with IP\cite{zhang2021event} } & { ANN2SNN } & { 98.45 }\\
  \textsf{  } & {BackEISNN\cite{zhao2022backeisnn}  } & { directed trained } & { 99.67 }\\
  \textsf{  } & {Ling et al.\cite{liang2021exploring}  } & { spike-based BP } & { 99.52 }\\
  \textsf{ MNIST } & {BRP-SNN\cite{zhang2021tuning}  } & { spike-based BRP } & { 99.01 }\\
  \textsf{  } & {STDBP\cite{zhang2021rectified} } & { spike-based BP } & { 99.4 }\\
  \textsf{  } & {Zhu et al.\cite{zhu2022training}} & { Time-based BP} & { 99.47 }\\
  \textsf{  } & {CL-SNN with fixed $\epsilon$(\textbf{ours})  } & { Spike-based BP } & { \textbf{99.71} }\\
  \textsf{  } & {CL-SNN with dynamic $\epsilon$(\textbf{ours})  } & { Spike-based BP } & { \textbf{99.68} }\\
\specialrule{0.1em}{3pt}{3pt} 
  \textsf{  } & {SCNN with IP\cite{zhang2021event}  } & { ANN2SNN } & { 81.65 }\\
  \textsf{  } & {ANN with full-precision activation\cite{wu2021tandem}  } & { Error Backproagation } & { 91.77 }\\
  \textsf{  } & {BackEISNN\cite{zhao2022backeisnn}  } & { directed trained } & { 90.93 }\\
  \textsf{  } & {Ling et al.\cite{liang2021exploring}  } & { spike-based BP } & { 77.27 }\\
  \textsf{  CIFAR10} & {BRP-SNN\cite{zhang2021tuning}  } & { spike-based BRP } & { 57.08 }\\
  \textsf{  } & {Wu et al.\cite{wu2019direct}  } & {STBP with NeuNorm } & { 90.53 }\\
  \textsf{  } & {Zhu et al.\cite{zhu2022training}} & { Time-based BP} & { 92.45 }\\
  \textsf{  } & {CL-SNN with fixed $\epsilon$(\textbf{ours}) } & { Spike-based BP } & { \textbf{92.74} }\\
  \textsf{  } & {CL-SNN with dynamic $\epsilon$(\textbf{ours})  } & { Spike-based BP } & { \textbf{93.03} }\\
  \specialrule{0.1em}{3pt}{0pt}
  \textsf{  } & {SCNN with IP\cite{zhang2021event}  } & { ANN2SNN } & { 92.62 }\\
  \textsf{  } & {BackEISNN\cite{zhao2022backeisnn}  } & { directed trained } & { 93.45 }\\
  \textsf{ Fashion-MNIST } & {STDBP\cite{zhang2021rectified}  } & { spike-based BP } & { 90.1 }\\
  \textsf{  } & {Zhu et al.\cite{zhu2022training}} & { Time-based BP}  & { 93.28 }\\
  \textsf{  } & {CL-SNN with fixed $\epsilon$(\textbf{ours})  } & { Spike-based BP } & {\textbf{94.54} }\\
  \textsf{  } & {CL-SNN with dynamic $\epsilon$(\textbf{ours})  } & { Spike-based BP } & { \textbf{94.38} }\\
\specialrule{0.1em}{3pt}{3pt}
\end{tabular}}
\end{center}
\end{table*}

\begin{table*}
\begin{center}
\caption{Comparisions of accuracy with other methods on neuromorphic datasets.}
\label{tab:results2} \scalebox{1.0}{
\begin{tabular}{cccc}
\specialrule{0.1em}{5pt}{3pt} 
\textsf{ \bf{ Dataset }} & \textbf{ model   } & \textbf{ method   } & \textbf{ accuracy     }  \\
\specialrule{0.1em}{3pt}{0pt}\\
  \textsf{  } & {BackEISNN\cite{zhao2022backeisnn}  } & { directed trained } & { 99.57 }\\
  \textsf{  } & {Ling et al.\cite{liang2021exploring}  } & { spike-based BP } & { 99.49 }\\
  \textsf{ N-MNIST } &  {Wu et al.\cite{wu2019direct}  } & {STBP with NeuNorm }  & { 99.53 }\\
  \textsf{  } &  {Zhu et al.\cite{zhu2022training}} & { Time-based BP} & { 99.39 }\\
  \textsf{  } & {CL-SNN with fixed $\epsilon$(\textbf{ours})  } & { { Spike-based BP }  } & { \textbf{99.58} }\\
  \textsf{  } & {CL-SNN with dynamic $\epsilon$(\textbf{ours})  } & { Spike-based BP } & { \textbf{99.63} }\\
\specialrule{0.1em}{3pt}{3pt} 
  \textsf{  } & {Ling et al.\cite{liang2021exploring}  } & { spike-based BP } & { 64.6 }\\
  \textsf{  CIFAR10-DVS} & {Hanle et al.\cite{zheng2021going}  } & { STBP-tdBN }  & { 67.8 }\\
  \textsf{  } &  {Wu et al.\cite{wu2019direct}  } & {STBP with NeuNorm }  & { 60.5 }\\
  \textsf{  } & {CL-SNN with fixed $\epsilon$(\textbf{ours})  } & {{ Spike-based BP }  } & {\textbf{69.4}  }\\
  \textsf{  } & {CL-SNN with dynamic $\epsilon$(\textbf{ours})  } & { Spike-based BP } & { \textbf{68.6} }\\
  \specialrule{0.1em}{3pt}{0pt}
  \textsf{  } & {Ling et al.\cite{liang2021exploring} } & { spike-based BP } & { 91.32 }\\
  \textsf{ DVS-Gesture } & {BRP-SNN\cite{zhang2021tuning}  } & { spike-based BRP }& { 80.9 }\\
  \textsf{  } & {CL-SNN with fixed $\epsilon$(\textbf{ours})  } & { Spike-based BP } & {\textbf{ 94.44} }\\
  \textsf{  } & {CL-SNN with dynamic $\epsilon$(\textbf{ours})  } & { Spike-based BP } & { \textbf{94.72} }\\
\specialrule{0.1em}{3pt}{3pt}
\end{tabular}}
\end{center}
\end{table*}



\subsection{Other metrics}
We recorded other evaluation metrics of the model on six datasets, including macro Precision, macro Recall, and macro F1-score, hoping to provide some assistance for future research. 

Macro Precision is a metric used in multi-class classification tasks to measure the average precision across all classes and provides an overall measure of the precision performance across the different classes in the classification task.
Similar to macro Precision, macro Recall is a metric used in multi-class classification tasks to measure the average recall across all classes. It calculates the average of the recall values for each class.
Macro F1-score provides an overall measure of the F1-score performance across the different classes in the classification task.The macro Precision, macro Recall, and macro F1-score for used datasets are show in Table\eqref{tab:metrics}.
\begin{table*}
\begin{center}
\caption{\centering{The proposed CL-SNN model's macroPrecision, macroRecall, macroF1-score on diffierent datasets}}
\resizebox{\linewidth}{!}{
\label{tab:metrics} \scalebox{0.6}{
\begin{tabular}{ccccccc}
\specialrule{0.1em}{3pt}{3pt} 
 \textsf{ \bf{ Dataset }} & \textbf{ MNIST   } & \textbf{ CIFAR10   } & \textbf{ Fashion-MNIST  }& \textbf{ N-MNIST   } & \textbf{ CIFAR10-DVS   } & \textbf{ DVS-Gesture  }   \\[1ex]
  \textsf{ macroPrecision } & { 0.9971  } & { 0.9275 } & { 0.9451 }& {0.9958  } & { 0.6873 } & { 0.9446 }\\[1ex]
  \textsf{ macroRecall } & {0.9970  } & { 0.9274 } & { 0.9454 }& {0.9957  } & { 0.6870 } & { 0.9413 }\\[1ex]
  \textsf{ macroF1score } & {0.9970  } & { 0.9274 } & { 0.9957 }& { 0.6852 } & { 0.9275 } & { 0.9406 }\\
\specialrule{0.1em}{3pt}{3pt}
\end{tabular}}}
\end{center}
\end{table*}

\section{Conclusion and disscusion}
The previous SNNs model always processed all samples indiscriminately, which was not in line with the natural process of human learning new knowledge from easy to difficult. We propose to introduce CL based on confidence perception loss function into SNNs. CL is a training strategy that makes the model learn knowledge from easy to difficult. It is inspired by the process of human learning new knowledge. And based on this, a CL-SNN model was proposed, which has a high degree of biological rationality. By scaling the contributions of samples with different confidence levels in parameter updates, the core principles of course learning are achieved. To our knowledge, this is the first proposal to enhance the biological rationality of SNNs by introducing CL.

However, determining the most suitable curriculum learning strategy for a specific task still requires further exploration. Confidence-aware loss is a convenient method, but the choice of difficulty threshold and the measurement of difficulty are not unique. Nonetheless, this paper provides insights into how to design SNNs that align with human cognitive processes.

\section*{Acknowledgments}
This work is supported by the National Natural Science Foundation of China (No. 62276218), Sichuan Science and Technology Program (No. 2022YFG0031) and Chengdu International Science and Technology Cooperation (No. 2023-GH02-00029-HZ).

\bibliographystyle{IEEEtran}
\bibliography{CL-SNN}

\vfill

\end{document}